\documentclass[conference]{IEEEtran}
\IEEEoverridecommandlockouts
\usepackage{cite}
\usepackage{amsthm}
\usepackage{amsmath,amssymb,amsfonts}
\usepackage{algorithmic}
\usepackage{graphicx}
\usepackage{textcomp}
\usepackage{xcolor}
\usepackage{cleveref}

\usepackage{enumitem}
\usepackage{etoolbox}
\usepackage{graphicx}
\usepackage{makecell}
\usepackage{caption}
\usepackage{wrapfig}
\usepackage{setspace}
\usepackage{subcaption}
\usepackage{floatflt}
\usepackage{float}
\usepackage{setspace}

\newtheorem{theorem}{Theorem}

\DeclareMathOperator*{\argmax}{argmax}

\renewrobustcmd{\bfseries}{\fontseries{b}\selectfont}
\renewrobustcmd{\boldmath}{}
\newrobustcmd{\B}{\bfseries}

\newsavebox\CBox

\makeatletter

\Crefname{equation}{Eq.}{Eqs.}
\Crefname{figure}{Fig.}{Figs.}
\Crefname{tabular}{Tab.}{Tabs.}
\Crefname{table}{Tab.}{Tabs.}
\Crefname{section}{Sec.}{Sec.}

\def\BibTeX{{\rm B\kern-.05em{\sc i\kern-.025em b}\kern-.08em
    T\kern-.1667em\lower.7ex\hbox{E}\kern-.125emX}}
\begin{document}

\title{Classification Error Bound for Low Bayes Error Conditions in Machine Learning
% \\
% {\footnotesize \textsuperscript{*}Note: Sub-titles are not captured for https://ieeexplore.ieee.org  and
}
\author{
    \IEEEauthorblockN{Zijian Yang\IEEEauthorrefmark{1}\IEEEauthorrefmark{2}, Vahe Eminyan\IEEEauthorrefmark{1}, Ralf Schlüter\IEEEauthorrefmark{1}\IEEEauthorrefmark{2}, Hermann Ney\IEEEauthorrefmark{1}\IEEEauthorrefmark{2}}
    \IEEEauthorblockA{\IEEEauthorrefmark{1}Machine Learning and Human Language Technology Group, Lehrstuhl Informatik 6,\\
Computer Science Department, RWTH Aachen University, Germany}
    \IEEEauthorblockA{\IEEEauthorrefmark{2}AppTek GmbH, Germany}
}

% \and
% \and
% \and
% \and
% \and

\maketitle

\begin{abstract}
    
    % The error mismatch problem is a challenge for statistical classification systems. The problem is introduced in the decision rule by replacing the true distributions with model distributions estimated on training data. To investigate the mismatch condition, classification error bounds

% We study the classification error of a machine learning system by strictly distinguishing between the probabilistic model learned in training and the true distribution as used in Bayes decision rule. This is referred to as the error mismatch condition.
% The classification error bound is applied to study the relationship between the error mismatch and statistical measures.
% Motivated by the observation that the Bayes error is typically low in machine learning tasks like speech recognition, tight bound on the error mismatch based on Kullback–Leibler divergence with the constraint that the Bayes error is lower than a threshold. We show that the bound can be approximated linearly. The error bound is further extended to the case with strings. With such extension, the correlations between the cross-entropy loss, language model perplexity, and word error rate are discussed analytically.
% In statistical classification and machine learning, the unknown true distribution is usually replaced with a model distribution estimated from the training data in the Bayes decision rule. 
In statistical classification and machine learning, classification error is an important performance measure, which is minimized by the Bayes decision rule. In practice, the unknown true distribution is usually replaced with a model distribution estimated from the training data in the Bayes decision rule. 
This substitution introduces a mismatch between the Bayes error and the model-based classification error. In this work, we apply classification error bounds to study the relationship between the error mismatch and the Kullback-Leibler divergence in machine learning. Motivated by recent observations of low model-based classification errors in many machine learning tasks, bounding the Bayes error to be lower, we propose a linear approximation of the classification error bound for low Bayes error conditions. Then, the bound for class priors are discussed. Moreover, we extend the classification error bound for sequences. Using automatic speech recognition as a representative example of machine learning applications, this work analytically discusses the correlations among different performance measures with extended bounds, including cross-entropy loss, language model perplexity, and word error rate.

\end{abstract}

\section{Introduction}
In statistical classification and machine learning tasks, such as automatic speech recognition (ASR), Bayes decision rule is used to minimize the classification error, which is the most critical performance measure for these tasks.
However, since the true distribution in Bayes decision rule is unknown, in practice, a probabilistic model trained on the training data is applied to approximate the true distribution in Bayes decision rule.
Thus, there is a difference between the true distribution of the data and the probabilistic model \cite{ney2003relationship,schluter2013novel}. While  this difference is not addressed in most of the studies, we will make a mathematically strict distinction between true and model distributions in this work. 
% We refer to this distinction as the mismatch condition, because there is typically
% a mismatch between the true and the model distribution. 
As the classification error mismatch between the Bayes error and the model-based decision error reflects the proximity of model performance to the optimum, we will study the relationship between error mismatch and other statistical measures in this work.
% As the Bayes error is fixed for a given task, minimizing the error mismatch effectively leads to the improvement of the model performance.

Kullback–Leibler (KL) divergence is another important statistical measure in machine learning tasks. It is closely associated with the cross-entropy (CE) loss and language model (LM) perplexity (PPL).
While the correlation between word error rate (WER) and LM PPL has been observed for a long time \cite{bahl1983maximum, makhoul1995state, klakow2002testing}, most of the works only empirically demonstrate the correlation. In this work, we aim to examine the relationship from an analytical perspective. 
% To the best of our knowledge, this is the first theoretical investigation of this correlation.

The relationship between error mismatch and KL divergence was first investigated in \cite{ney2003relationship}. There, Ney introduced two error bounds on the error mismatch. Later, Nussbaum et al. derived a generalized statistical bound on error mismatch \cite{schluter2013novel, nussbaum2013relative}, which included the KL divergence as an implicit upper bound of the error mismatch. The bound derived in \cite{schluter2013novel, nussbaum2013relative} was proven to be tight when the true distribution is arbitrary. However, when more information about the true distribution is obtained, the bound can be improved. In practice, many systems/tasks have low Bayes errors. For instance, the WER of human speech recognition, is typically low \cite{wesker2005oldenburg}, often dropping below 1\% for a wide range of conditions \cite{lippmann1997speech}, indicating a bound on the Bayes error to be lower. In \cite{yang2024refined}, a refined tight bound between error mismatch and KL divergence is derived when Bayes error is lower than a threshold. In this work, we will revisit classification error bounds within the context of machine learning under the low Bayes error condition. Contributions of this work are as follows:
\begin{itemize}
    \item Simplify the bound with a linear approximation under the low Bayes error condition
    \item Propose the bound for class priors and verify its tightness
    \item Extend classification error bounds for sequences
\end{itemize}
With the extended bounds, correlations among different performance measures including cross-entropy, language model perplexity and word error rate will be discussed.

% In this work, we will employ the bound derived in \cite{yang2024refined} to explore topics in machine learning, especially in the domain of ASR. Based on the assumption of the low Bayes error, we will present a linear approximation of the bound from \cite{yang2024refined} under this condition. To investigate the relationship between LM PPL and WER, the bound for class priors are then introduced. Considering sequence-to-sequence machine learning tasks, we extend the bound to accommodate sequences with a position-wise-defined error. As an application of extended bounds, correlations between CE loss/PPL and WER in the ASR domain will be discussed analytically.

% Given that the bound derived in \cite{yang2024refined} applies to single events and ASR involves sequence-to-sequence tasks, we will extend the bound to suit sequence cases. With the help of the linear bound, the correlation between CE loss and WER, as well as between PPL and WER will be discussed.
% To this end, we will first provide a linear approximation of the bound, and discuss the bound for class priors. To investigate the measures defined on the sequence level, we will extend the bound to accommodate strings with a position-wise-defined error. With the help of the linear bound, 

 % which is widely used for modern neural architectures such as connectionist temporal classification \cite{graves2006connectionist}, neural transducers \cite{graves2012sequence} and attention-based models \cite{bahdanau2016end, chan2016listen}.
% \vspace{-1.5mm}
\section{Statistical Measures}
In a statistical classification problem, also known as multiple hypothesis testing in information theory, for a joint event $(c,x)$, where $c\in \mathcal{C}$ is a class and $x \in \mathcal{X}$ is a discrete observation, the expected error is defined as:
\begin{equation}
    E[c|x] = \sum_{c'} pr(c'|x) \big(1- \delta(c, c')\big) = 1 - pr(c|x)
\end{equation}
where $\delta$ is Kroneker-Delta and $pr(c|x)$ is the true posterior distribution. The minimum classification error is obtained by the Bayes decision rule:
% Consider a statistical classification problem, where $pr(c,x)$ is defined as the joint true distribution for a class $c\in \mathcal{C}$ and an observation $x \in \mathcal{X}$. To simplify the derivation, we assume that $x$ is a discrete variable, while all the derivations can be extended to continuous variables by replacing the summation with the integral. The Bayse decision rule is defined as:
\begin{equation}
    c_*(x) =  \argmax_c pr(c|x)
    % \label{eq:bayes}
\end{equation}
In practical applications, the true distribution is unknown. Therefore, a model distribution $q(c,x)$ is employed to estimate the true distribution. Recently, sequence-to-sequence (Seq2Seq) modeling methods have achieved significant success in machine learning tasks, attaining low classification errors on different tasks. \cite{prabhavalkar2023end,graves2006connectionist,graves2012sequence, bahdanau2016end}. Instead of modeling the joint distribution $q(c,x)$, these models directly model the posterior $q(c|x)$. For generalization purposes, we consider the modeling of joint distribution $q(c,x)$ in this paper, unless otherwise specified. All the results can be generalized to the posterior modeling by defining $q(c,x):= q(c|x) \cdot pr(x)$. The model-based decision rule is defined as:
\begin{equation}
    c_q(x) = \argmax_c q(c|x)
    % \label{eq:localdecision}
\end{equation}

% \begin{figure}[htb]
%     \centering
%     \includegraphics[scale=0.3]{delta_general.png}
%         \caption{Nussbaum, Bretagnolle-Huber, Ney, Vajda bound on error mismatch, compared to simulations of pairs of distributions. No constraint on $E_*$ is applied. The grey dots refer to the simulation points.}
%             \label{fig:existing_bounds}
%             \vspace{-4mm}
% \end{figure}
% Since $pr(x)$ does not influence the decision rule, the modeling of $q(c,x)$ can be interpreted as $q(c,x) = q(c|x) pr(x)$. Unless otherwise stated, we will use this interpretation in the following discussions. Similarly, the model-based decision rule is defined as:\\
% \scalebox{0.85}{\parbox{1.05\linewidth}{
% \begin{equation}
%     c_q(x) =  \argmax_c q(c,x) = \argmax_c q(c|x)
%     \label{eq:localdecision}
% \end{equation}}}
% Based on the decision rules, the local Bayes and model-based classification errors, $E_*(x)$ and $E_q(x)$, as well as the local classification error mismatch $\Delta_q(x)$ are defined as:\\
% \scalebox{0.95}{\parbox{1.05\linewidth}{
% \begin{align}
%     E_*(x) &=1- pr(c_*(x)|x), \quad E_q(x) =1- pr(c_q(x)|x)\\
%     \Delta_q(x) &= E_q(x) - E_*(x)
% \end{align}
% }}
% It is worth to note that $0\leq \Delta_q(x) \leq 1$ according to the definition. The global Bayesian and model-based classification errors are then obtained by summing over all the observations.
In statistical classification, the most important performance criterion is the classification error. The global Bayesian and model-based classification errors are then obtained by computing the expectation across all observations:
\begin{equation}
\begin{aligned}
    E_* = \sum_{x} pr(x) E[c_*(x)|x], \quad E_q  = \sum_x pr(x) E[c_q(x)|x]
\end{aligned}
\end{equation}

In the mismatch problem, we are interested in the global classification error mismatch $\Delta_q$ between $E_*$ and $E_q$:\\
\scalebox{0.95}{\parbox{1.05\linewidth}{
\begin{equation}
    \Delta_q = E_q - E_* = \sum_x pr(x) \big (pr(c_*(x)|x) - pr(c_q(x)|x) \big)
\end{equation}
}}
Note that $\Delta_q \geq 0$, i.e. $E_q$ is lower bounded by $E_*$. Therefore, minimizing $\Delta_q$ pushes the model toward achieving the optimal classification error. 

KL divergence is another statistical measure used to assess the difference between two distributions, which is defined as:\\
\scalebox{0.95}{\parbox{1.05\linewidth}{
\begin{align}
    D_\text{KL}(pr \parallel  q) &= \sum_{x,c} pr(c,x) \log \frac{pr(c,x)}{q(c, x)} 
    % & =  \sum_x pr(x) \sum_c pr(c|x) \log \frac{pr(c|x)}{q(c|x)}\\
    % & = \sum_x pr(x) D_\text{KL}(pr(c|x)\parallel q(c|x))
\end{align}
}}

\section{Classification Error Bounds}
\subsection{Existing Error Bounds}
In the field of information theory, instead of the error mismatch $\Delta_q$, the relationship between total variation distance $V$ and KL divergence has been elucidated in the past years.
% \scalebox{0.85}{\parbox{1.15\linewidth}{
% \begin{equation}
%     V := \sum_{x,c}|pr(c,x) - q(c,x)| \geq \Delta_q.
%     \label{eq:total_variational}
% \end{equation}
% }}
In \cite{fedotov2003refinements}, Vadja proposed a refinement of \textit{Pinsker's} inequality.
% \scalebox{0.85}{\parbox{1.15\linewidth}{
% \begin{equation*}
%     D_\text{KL}(pr \parallel q) \geq \log\frac{2+V}{2-V} - \frac{2V}{2-V}
% \end{equation*}
% }}
In machine learning, \cite[p.10]{vapnik1998statistical} introduced the \textit{Bretagnolle-Huber} bound for density estimation.
% \scalebox{0.85}{\parbox{1.15\linewidth}{
% \begin{equation*}
%     V \leq 2 \sqrt{1- \text{exp}(-D_\text{KL}(pr \parallel q))}
% \end{equation*}}}
These bounds were not proposed for the error mismatch $\Delta_q$. However, as pointed out in \cite{ney2003relationship} that $V$ is lower bounded by $\Delta_q$, the bounds for $\Delta_q$ can be obtained by replacing the total variation distance $V$ in the inequalities with $\Delta_q$.
In \cite{ney2003relationship}, starting from $V$, Ney also derived an error bound for $D_\text{KL}(pr \parallel q)$ as a function of $\Delta_q$ more directly.
% \scalebox{0.85}{\parbox{1.15\linewidth}{
% \begin{equation*}
%     D_\text{KL}(pr \parallel q) \geq (\frac{\Delta_q}{\sqrt{2}})^2.
% \end{equation*}
% }}
Nevertheless, unlike $V$, the error mismatch $\Delta_q$ is asymmetric. As a result, introducing $V$ in the derivation leads to a non-tight bound for $\Delta_q \in (0,1]$. In \cite{schluter2013novel, nussbaum2013relative}, started directly from $\Delta_q$, the following bound of $\Delta_q$ on $D_\text{KL}(pr \parallel q)$ is derived, which is a tight bound for the entire domain of $\Delta_q$ when the true and model distributions are unconstrained.
\begin{equation}
\begin{aligned}
    D_\text{KL}(pr\parallel q) &\geq \frac{1}{2}\big ((1+\Delta_q) \log(1+ \Delta_q)\\
     & \quad  \quad + (1-\Delta_q) \log(1- \Delta_q) \big )\\
    &: = g(\Delta_q)
    \label{eq:globalbound}
\end{aligned} 
\end{equation}

% The tight bound $g$ can also be applied to the local KL divergence and local mismatch:
% % \scalebox{0.85}{\parbox{1.15\linewidth}{
% \begin{equation}
%     D_\text{KL}(pr(c|x)\parallel q(c|x)) \geq g(\Delta_q(x))
%      \label{eq:localbound}
% \end{equation}
% % }}
% which can be derived similarly as the global bound in \cite{schluter2013novel}.
% The proof can be done similarly to the global bound derived in \cite{schluter2013novel}

% Figure \ref{fig:existing_bounds} illustrates the comparison of existing bounds and simulation results, with each grey dot representing the result of a single simulation. The simulation was conducted by generating various distribution pairs $(pr,q)$ until all the reachable areas were covered. All the simulations in this paper are done with $|\mathcal{C}|=7$ and $|\mathcal{X}|=15$. As shown in the figure, when the true distribution is arbitrary, the bound derived in \cite{schluter2013novel,nussbaum2013relative} is tight.
% for $\Delta_q \in [0,1]$.

\subsection{Error bounds with Constraints on $E_*$}

The bound $g$ introduced in \eqref{eq:globalbound} is a tight bound when the true distribution $pr$ is unconstrained. However, if the true distribution is subject to some constraints, the bound can be further improved. 
In machine learning tasks like pattern and speech recognition, the Bayes error is typically low. For instance, WERs of human speech recognition can be below 1\% for a wide range of conditions \cite{lippmann1997speech}, and the Bayes error is bounded to be lower. Under the constraint that $E_*\leq t < 0.5$, where $t$ is a given threshold, the bound can be refined.

% In \cite{yang2024refined}, a refined bound under the constraint $E_*\leq t < 0.5$ is derived, where $t$ is a given threshold. For the sake of completeness, we present the result from \cite{yang2024refined} here.

% Motivated by this, we consider a system with low Bayes error $E_*\leq t < 0.5$, where $t$ is an estimated threshold. A refined tight bound under this condition is proposed in Theorem \ref{theorem:refinedbound}. When $t\geq 0.5$, the bound $g$ proposed in \cite{nussbaum2013relative, schluter2013novel} will still be tight for $\Delta_q \in [0,1]$, which will not be discussed here.

\begin{figure}[!htb]
    \centering
    \includegraphics[scale=0.25]{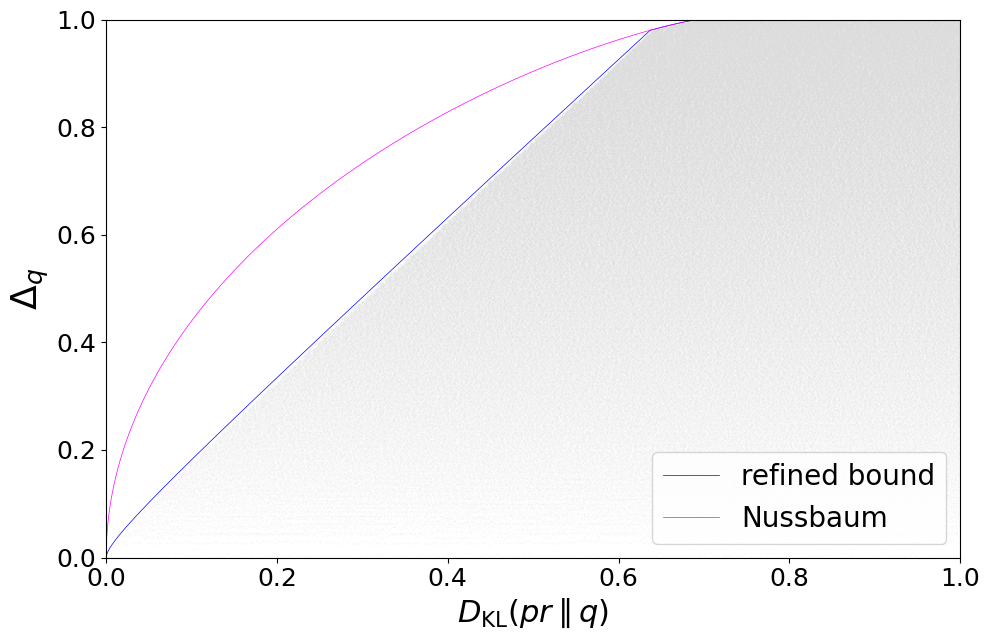}
    \caption{Comparison of the Nussbaum bound and the refined bound in Theorem \ref{theorem:refinedbound}. The simulation is done under the constraint $E_*\leq 0.01$. The grey dots refer to simulation points.}
    \label{fig:all_bounds}
\end{figure}
\begin{theorem}
\label{theorem:refinedbound}
    When $E_*\leq t< 0.5$, $D_\text{KL}(pr\parallel q)$ is lower-bounded by the following function of the mismatch $\Delta_q$,\\
    \scalebox{0.95}{\parbox{1.05\linewidth}{
    \begin{equation}
    \begin{aligned}
        D_\text{KL}(pr\parallel q) \geq \underbrace{
        \left \{ \begin{array}{lr}
     (\Delta_q+ 2t) g(\frac{\Delta_q}{\Delta_q+2t}),& 0 \leq \Delta_q < 1-2t\\
        g(\Delta_q), &1-2t \leq \Delta_q \leq 1 
    \end{array}
\right .}_{:=h_t(\Delta_q)}
\end{aligned}
\end{equation}
}}
where $h_t$ is the refined bound, and $g$ is defined as in \eqref{eq:globalbound}.
\end{theorem}

A detailed proof and the tightness of the bound are derived in our previous work \cite{yang2024refined}. Figure \ref{fig:all_bounds} shows the comparison of the bound derived in \cite{nussbaum2013relative, schluter2013novel} and the refined bound in \cite{yang2024refined},  with each grey dot representing the result of a single simulation. The simulation was conducted by generating various distribution pairs $(pr,q)$ until all the reachable areas were covered. All the simulations in this paper are done with $|\mathcal{C}|=7$ and $|\mathcal{X}|=15$. The simulation result show that the bound derived in \cite{nussbaum2013relative, schluter2013novel} is not tight under the constrained $E_*\leq t$, while the bound in \cite{yang2024refined} exhibits tightness across the full range of $\Delta_q$.

\begin{figure}[htb]
  \centering
  \includegraphics[height=4cm]{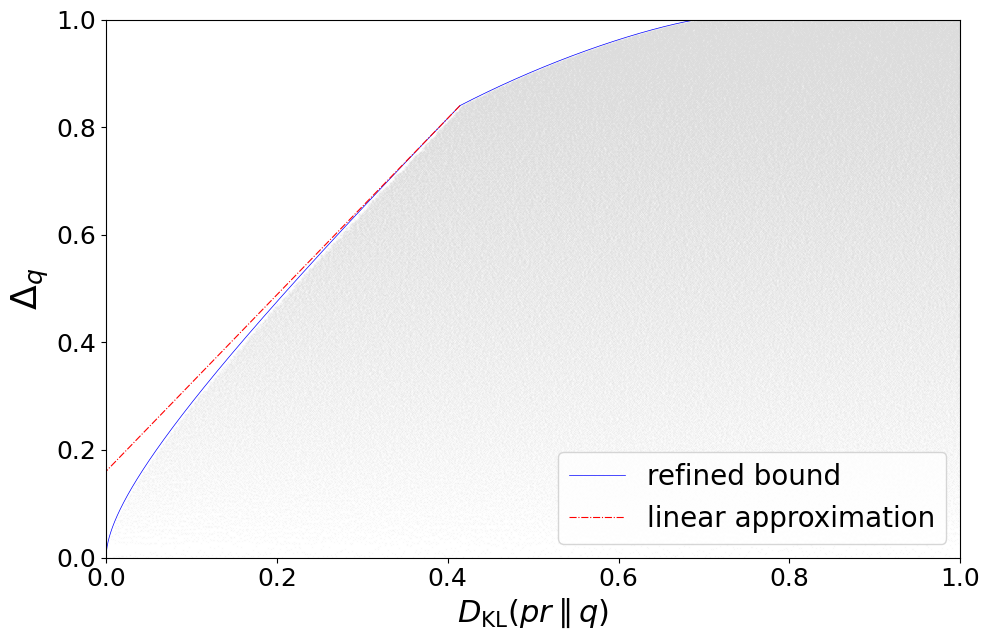}
  \includegraphics[height=4cm]{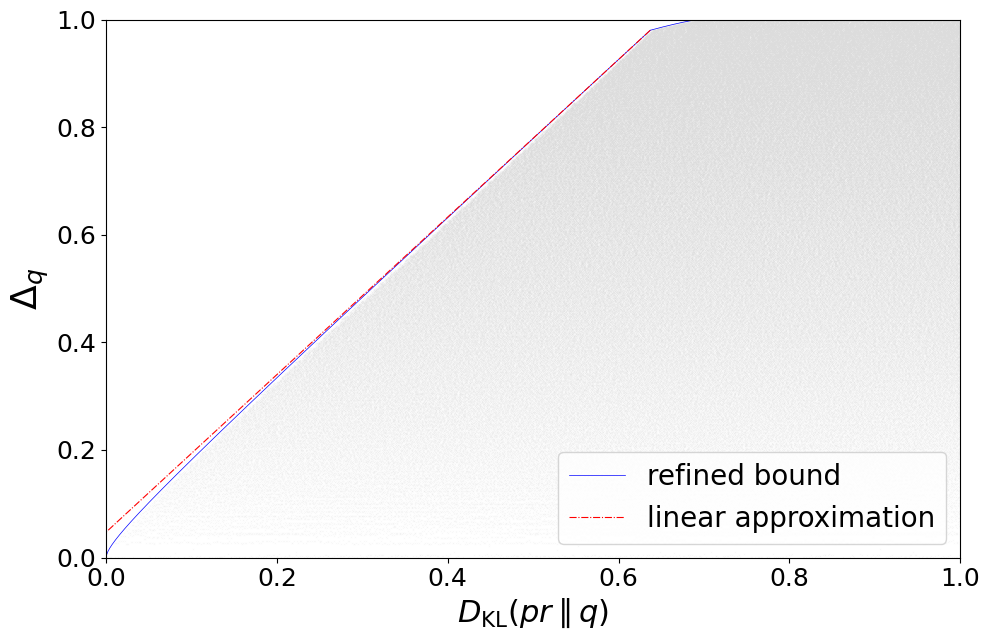}
  \caption{The linear approximation of the refined bound in Theorem \ref{theorem:refinedbound}. The simulations in the upper figure are under the constraint $E_* \leq 0.08$, and for the lower figure, the constraint is $E_* \leq 0.01$. Grey dots refer to the simulation points.}
  \label{fig:linear}
  \vspace{-5mm}
\end{figure}
\subsection{Linear Approximation of the Bound}

By computing the derivative of $h_t(\Delta_q)$, it can be observed that when $\Delta_q \gg t$ and within the range $0\leq {\Delta}_q \leq 1-2t$, the derivative is almost a constant value. Therefore, the bound can be approximated linearly when $t$ is small:
% \scalebox{0.85}{\parbox{1.15\linewidth}{
\begin{align}
    D_\text{KL}(pr \parallel q) \geq  \log(2-2t)  \cdot {\Delta}_q + \beta.
    \label{eq:linearapprox}
\end{align}
% }}
where $\beta=t \cdot  (\log(1-t) + \log t+2\log2)$. Note that since $h_t$ is convex, this linear bound is valid. Figure \ref{fig:linear} demonstrates the comparison between the refined bound and its linear approximation. As illustrated in the lower figure for $t=0.01$, the exact bound is effectively approximated by the linear bound across nearly the entire domain of $\Delta_q$. However, for $t=0.08$ in the upper figure, the accuracy of the approximation diminishes when $\Delta_q$ is small, as $\Delta_q \gg t$ is not fulfilled.
% the linear bound is a good approximation of the exact bound.

\subsection{Error Bound for Class Priors}
\label{sec:classprior}
Class prior is important in many statistical classification tasks. For instance, in ASR, an LM (class sequence prior) is usually combined with the acoustic model (AM) to achieve better performance. Therefore, it is crucial to investigate the effect of the class prior. 
% In ASR tasks, an LM is usually combined with the acoustic model (AM) to achieve better performance. Therefore, it is crucial to investigate the effect of the LM, namely, the class prior. 
To quantify the discrepancy between the true class prior and the model class prior, the KL divergence between two priors $pr(c)$ and $q(c)$ is employed.
\begin{equation*}
    D_\text{KL}\big(pr(c) \parallel q(c)\big) = \sum_c pr(c) \log \frac{pr(c)}{q(c)}
\end{equation*}
To eliminate the influence of modeling the AM, we assume a perfect acoustic model $q'(x|c) = pr(x|c)$. Effectively, we apply such modeling $q'(c, x) =q'(c) pr(x|c)$ where $q'(c)$ is the model prior for classes. In this case, the KL divergence between joint distributions collapses to between class priors, and the bound proposed in Theorem \ref{theorem:refinedbound} can be applied:

\begin{equation}
     D_\text{KL}(pr \parallel q') = \sum_c pr(c) \log \frac{pr(c)}{q'(c)} \geq h_t(\Delta_{q'})
\end{equation}
Due to the specific assumption of the joint model distribution $q'(c,x)$, we must reconsider the tightness of the bound.
The equality for $\Delta_{q'} \in [0,1-2t)$ can be achieved with the following parameterized distribution with parameter $\lambda \in [0.5, 1-2t)$:
\allowdisplaybreaks
\begin{align}
    pr(c, x_1) &= \left \{\begin{array}{ll}
         1- \frac{t}{1-\lambda}, &  c= c_1\\
         0, &\text{otherwise} 
    \end{array}
    \right .\\
   pr(c, x_2) &=   \left \{\begin{array}{ll}
         \frac{t\lambda}{1-\lambda}, & \quad c= c_2\\
         t, & \quad c=c_3\\
         0, &\quad \text{otherwise} 
         \end{array}     \right .\\
    q'(c) &= \lim_{\epsilon \rightarrow 0^+}\left \{\begin{array}{ll}
    1-\frac{t}{1-\lambda}, & c = c_1\\
    \frac{t}{1-\lambda} \cdot (0.5 - \epsilon), & c = c_2 \\
    \frac{t}{1-\lambda} \cdot (0.5 + \epsilon), & c = c_3 \\
    0, &\text{otherwise}
    \end{array} \right .&
\end{align}

\begin{figure}[htb]
    \centering
    \includegraphics[scale=0.25]{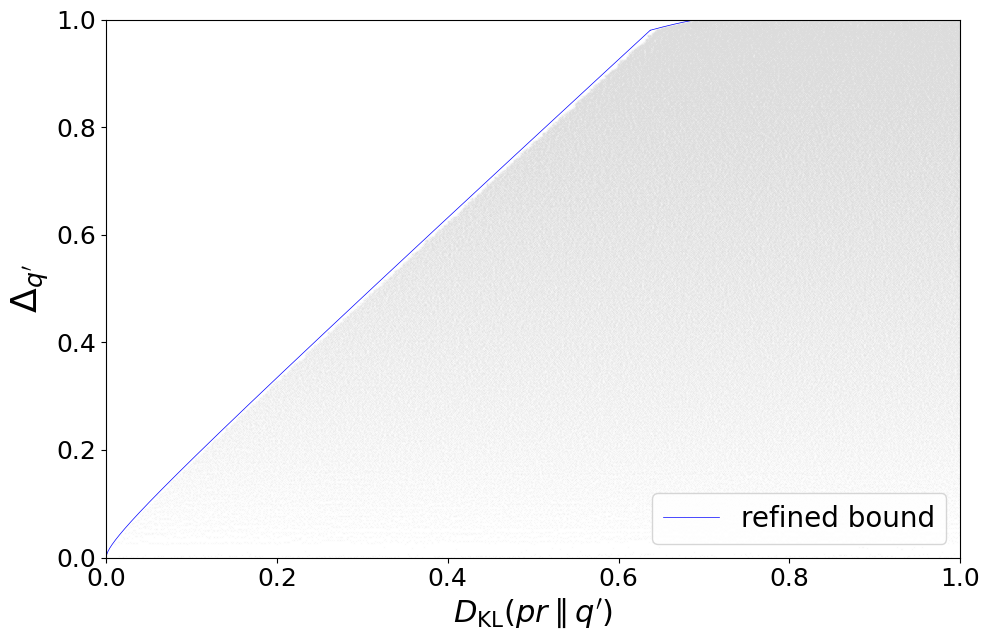}
    \caption{Simulation results for KL divergence between class priors vs. error mismatch. The simulation is done with $E_* \leq 0.01$. Grey dots refer to the simulation points.}
    \label{fig:prior}
    \vspace{-3mm}
\end{figure}

For $\Delta_{q'} \in [1-2t,1]$, since the distributions used to achieve equality in \cite{schluter2013novel} meet the condition $q(x|c) = pr(x|c)$, the same distributions can be applied to obtain equality here. Figure \ref{fig:prior} presents the simulation result for the KL divergence between class priors and the error mismatch. As shown in the figure, the bound derived for joint distributions also holds for class priors and maintains its tightness.
\section{Error Bound for Sequences}
Since many machine learning tasks involve Seq2Seq modeling, in this section, we will delve into the application of previously derived bounds within the context of sequence-related scenarios, which bridges the theoretical bound with the practical Seq2Seq machine learning tasks. To simplify the discussion, we assume that all the class sequences have the same length $N$. The class sequence is defined as $c_1^N$, while the observation sequence is defined as $X$.
A straightforward way to extend the results from single observations to sequences is to treat the full sequences $c_1^N$ and $X$ as individual events. In this case,  $\Delta_q$ is computed on sequence level, i.e. sentence error mismatch. However, in Seq2Seq machine learning tasks like ASR, metrics are typically defined at the token or state level for each position. Therefore, a position-wise-defined error function is needed. 
We consider a position-wise defined error function $L$ for the sequence pair $c_1^N$ and $\Tilde{c}_1^N$.\\
\scalebox{0.95}{\parbox{1.05\linewidth}{
\begin{align}
    L[c_1^N, \Tilde{c}_1^N] := \frac{1}{N}\sum_{n=1}^N [1 - \delta(c_n, \Tilde{c}_n)]
\end{align}
}}
The expected error for a given sequence pair $(X, c_1^N)$ is:\\
\scalebox{0.95}{\parbox{1.05\linewidth}{
\begin{align}
    E[c_1^N|X] &= \sum_{\Tilde{c}_1^N} pr(\Tilde{c}_1^N|X)L[c_1^N, \Tilde{c}_1^N] = 1 - \frac{1}{N} \sum_n pr_n(c_n|X)
    % = 1 - \sum_n \sum_{\Tilde{c}_1^N} \frac{q(c_1^N|X)\delta(c_n, \Tilde{c}_n)}{N}\\
\end{align}
}}
where $pr_n(c|X)$ is the marginal distribution at position $n$.
\begin{equation}
    pr_n(c|X) = \sum_{c_1^N: c_n=c} pr(c_1^N|X)
\end{equation}

For each position $n$, the minimum of the expected error is obtained by the following Bayes decision rule:
\begin{equation}
    c_*^n(X) = \argmax_c pr_n(c|X)
    % c_q^n(X) = \argmax_c q_n(c|X)
\end{equation}
Consequently, the Bayes decision rule and the corresponding Bayes error for the whole sequence are defined as:
\begin{align}
    \mathbf{c}_*(X) = c_1^N | c_n = c_*^n(X), \overline{E_*} = \sum_X pr(X) E[\mathbf{c}_*(X)|X] 
    % &= 1 - \sum_X pr(X) \frac{1}{N}\sum_n pr_n(c_*^n(X)|X)
\end{align}
The model-based decision rule and classification error can be defined similarly:\\
\scalebox{0.95}{\parbox{1.05\linewidth}{
\begin{align}
    c_q^n(X) &= \argmax_c q_n(c|X), \text{  } \mathbf{c}_q(X) = c_1^N | c_n = c_q^n(X) \\
    \overline{E_q} &= \sum_X pr(X) E[\mathbf{c}_q(X)|X]
    % &= 1 - \sum_X pr(X) \frac{1}{N}\sum_n pr_n(c_q^n(X)|X)
\end{align}
}}
The error mismatch for the whole sequence is then computed as follows:
\allowdisplaybreaks
\begin{equation}
        \overline{\Delta_q} = \overline{E_q} - \overline{E_*} = \frac{1}{N} \sum_n \Delta_q^n
\end{equation}
where
\begin{align}
    \Delta_q^n = \sum_{X} pr(X) \Big [pr_n\big(c_*^n(X)|X\big) - pr_n\big(c_q^n(X)|X\big) \Big]
\end{align}

By employing Ineq. (25) in \cite{yang2024refined} and log-sum inequality, the relationship between the sequence-level KL divergence $D_\text{KL}(pr \parallel q)$ and $\overline{\Delta_q}$ can be derived as follows:
\scalebox{0.95}{\parbox{1.05\linewidth}{
\allowdisplaybreaks
\begin{align}
&\underbrace{D_\text{KL}(pr \parallel q) \geq \sum_{X} pr(X) \sum_{c_1^N} pr(c_1^N|X) \log \frac{pr(c_1^N|X)}{q(c_1^N|X)}}_{\text{Ineq. (25) in \cite{yang2024refined}}} \notag \\
& = \sum_{X} pr(X) \frac{1}{N} \sum_n \sum_c \underbrace{\sum_{c_1^N: c_n=c}pr(c_1^N|X) \log \frac{pr(c_1^N|X)}{q(c_1^N|X)}}_{\text{apply log-sum inequality}} \notag \\
&\geq \sum_{X} pr(X) \sum_c \frac{1}{N} \sum_n  pr_n(c|X) \log \frac{pr_n(c|X)}{q_n(c|X)} \notag \\
& = \frac{1}{N} \sum_n \underbrace{\sum_{X} pr(X) \sum_c pr_n(c|X) \log \frac{pr_n(c|X)}{q_n(c|X)}}_{\text{apply Theorem \ref{theorem:refinedbound}}} \notag \\
&\geq \underbrace{\frac{1}{N} \sum_n h_t(\Delta_q^n)}_{h_t \text{ is convex}} \geq h_t(\frac{1}{N} \sum_n\Delta_q^n) = h_t(\overline{\Delta_q}) \label{eq:klvsavgerror}
% & \Rightarrow     D_\text{KL}(pr \parallel q) \geq h_t \big (\overline{\Delta_q} \big) \label{eq:klvsavgerror}
\end{align}
}}

\subsection{Error Bound for the Model Classification Error}
% \scalebox{0.85}{\parbox{1.15\linewidth}{%
% \begin{align*}
%     &D_\text{KL}(pr\parallel q) = \sum_{X} pr(X) \sum_{c_1^N} pr(c_1^N|X) \log \frac{pr(c_1^N|X)}{q(c_1^N|X)} \\
%     & = \sum_{X} pr(X) \frac{1}{N}\sum_n \sum_c \sum_{c_1^N:c_n=c} pr(c_1^N|X) \log \frac{pr(c_1^N|X)}{q(c_1^N|X)} \\
%     & \geq \sum_{X} pr(X) \frac{1}{N}\sum_n \sum_c pr_n(c|X) \log \frac{pr_n(c|X)}{q_n(c|X)}\\
%     & \geq \sum_{X} pr(X)  \frac{1}{N}\sum_n  g(\Delta_{qn}(X))\\
%     & \geq g(\sum_{X} pr(X)  \frac{1}{N}\sum_n  \Delta_{qn}(X)) \\
%     & = g(\Bar{\Delta}_q)
%     % & \geq \sum_{X} pr(X) \frac{1}{N}\sum_n \sum_{c_1^N:c_n=c} pr(c_1^N|X) \log \frac{\sum_{{c'}_1^N: c'_n=c}pr(c'_1^N|X)}{q(c_1^N|X)} 
% \end{align*}}}
 KL-divergence can be reformulated in terms of cross-entropy $H(pr,q)$ and entropy $H(pr)$.
\begin{equation}
\begin{aligned}
    D_\text{KL}(pr \parallel q) &= H(pr,q) - H(pr) 
\end{aligned}
\end{equation}

For a given task, the true distribution,  $\overline{E_*}$ and $H(pr)$ are fixed. Therefore, by applying \eqref{eq:linearapprox} and \eqref{eq:klvsavgerror}, the model error $\overline{E_q}$ is linear bounded by $H(pr,q)$:
\allowdisplaybreaks
\begin{align}
    D_\text{KL}(pr \parallel q)& \geq \log(2-2t)  \cdot (\overline{E_q}-\overline{E_*}) + \beta \notag \\
   \Rightarrow H(pr,q) &\geq \log(2-2t)  \cdot \overline{E_q} + \text{const}
    \label{eq:celinear}
\end{align}

% As discussed in \cite{ney2003relationship, nussbaum2014family}, 
\subsection{Error Bound and CE Training Loss}
When there is enough data, as discussed in \cite{ney2003relationship}, the true distribution can be approximated by the empirical distribution:
\begin{align}
    pr(c_1^N, X) &\approx \frac{1}{M} \sum_{m=1}^M \delta(c_1^N, \mathbf{c}_m) \cdot \delta(X, X_m),\\
    H(pr,q) &= -\sum_{X, c_1^N} pr(c_1^N,X) \log q(c_1^N, X) \notag \\
    &\approx -\frac{1}{M}\sum_{m=1}^M \log q(\mathbf{c}_m,X_m)
\end{align}
where $(\mathbf{c}_m, X_m)$ are sequence pairs in the training data and $M$ is the number of sequences. Substituting the true distribution with the empirical distribution transforms $H(pr,q)$ into the standard CE training loss. The error bound \eqref{eq:celinear} implies that the model error is linearly upper bounded by the CE loss.
% that reducing the CE loss correlates with a decrease in model error. 

% Compared to the bound proposed in \cite{ney2003relationship}, the linear bound (\ref{eq:celinear}) offers a more direct justification of the connection between the training loss and the position-wise-defined evaluation metric applied in ASR tasks.
\subsection{The Correlation between Word Error Rate and Perplexity}
In this section, we investigate the correlation between WER and LM PPL via the derived error bound. Since the exact WER computation involves the alignment problem, which makes the problem much more complicated, we study the averaged Hamming distance instead, which is an upper bound to the WER, if the hypothesis is not longer than the reference. By definition, $\overline{E_q}$ is the error rate when applying Hamming distance as the metric. Similar to the discussion in Section \ref{sec:classprior}, we assume a perfect acoustic model to eliminate the influence of modeling the AM. In this case, the cross-entropy is $H\big(pr(c_1^N), q'(c_1^N)\big)$, which is effectively the logarithm of the LM PPL.
% \begin{equation}
%     \begin{aligned}
%         &H\big(pr(c_1^N), q'(c_1^N)\big) = -\sum_{c_1^N} pr(c_1^N) \log q'(c_1^N)\\
%         &\approx \frac{1}{M} \sum_{m=1}^M \log q'(\mathbf{c}_m) = \frac{1}{M} \sum_{m=1}^M \sum_{n=1}^N \log q'(c_{n,m} |c_{1,m}^{n-1})\\
%         & = 
%     \end{aligned}
% \end{equation}
By applying \eqref{eq:celinear}, the relationship between LM PPL and WER can be approximately derived as: 
\begin{equation}
    \log \text{PPL} \geq \log(2-2t) \Bar{E}_{q'} + \text{const} \geq \log(2-2t) \text{WER} + \text{const} \notag
\end{equation}
This inequality indicates that the WER is linearly upper-bounded by the logarithm of PPL. 
% However, the exact relationship between WER and PPL remains to be elusive. Without further constraints on the model distribution, the lower bound on WER cannot be obtained. 
% Detailed constraints on the true and model distribution by the specific task in the real world might result in a more precise upper bound, and possibly identify a lower bound. 
% These bounds could provide a more accurate characterization of the relationship. 
In \cite{klakow2002testing}, a log-linear relationship is observed between PPL and WER. To verify this relationship in theory, refined lower/upper bounds with further constraints on true/model distributions would be needed.

% \begin{figure}
%     \centering
%     \includegraphics[scale=0.3]{general_KL_fixed_log_0999.png}
%     \caption{KL divergence between posteriors}
%     \label{fig:enter-label}
% \end{figure}

% \begin{figure}
%     \centering
%     \includegraphics[scale=0.3]{LM_cond_log_099.png}
%     \caption{KL divergence between priors}
%     \label{fig:enter-label}
% \end{figure}
\section{Conclusion}
% In this work, we proposed and proved a novel tight bound on the mismatch between Bayes and model-based classification error based on Kullback–Leibler divergence under a constraint that the Bayes error is lower than a threshold. The bound was validated by simulations. 
In this work, bounds on the mismatch between the Bayes and model classification error based on Kullback–Leibler divergence were discussed for low Bayes error conditions. A linear approximation of the bound was proposed for these low Bayes error conditions. Following discussions on the bound for class priors, classification error bounds were extended for sequences. Based on extended bounds, linear bounds between different performance metrics including cross-entropy loss, language model perplexity, and word error rate were derived.

\bibliographystyle{IEEEtran}
\bibliography{mybib}

\newpage

\end{document}